\documentclass[letterpaper]{article} % DO NOT CHANGE THIS
\usepackage{aaai24}  % DO NOT CHANGE THIS
\usepackage{times}  % DO NOT CHANGE THIS
\usepackage{helvet}  % DO NOT CHANGE THIS
\usepackage{courier}  % DO NOT CHANGE THIS
\usepackage[hyphens]{url}  % DO NOT CHANGE THIS
\usepackage{graphicx} % DO NOT CHANGE THIS
\usepackage{natbib}  % DO NOT CHANGE THIS AND DO NOT ADD ANY OPTIONS TO IT
\usepackage{caption} % DO NOT CHANGE THIS AND DO NOT ADD ANY OPTIONS TO IT
\usepackage{algorithm}
\usepackage{algorithmic}

\usepackage{amsmath}
\usepackage{multirow}
\DeclareMathOperator*{\argmin}{argmin}
\usepackage{caption}
\usepackage{subcaption}
\usepackage{amsfonts}

\usepackage{newfloat}
\usepackage{listings}
\DeclareCaptionStyle{ruled}{labelfont=normalfont,labelsep=colon,strut=off} % DO NOT CHANGE THIS
\floatstyle{ruled}
\newfloat{listing}{tb}{lst}{}
\floatname{listing}{Listing}
\title{GINN-LP: A Growing Interpretable Neural Network for Discovering Multivariate Laurent Polynomial Equations}
\author{
    Nisal Ranasinghe\textsuperscript{\rm 1},
    Damith Senanayake\textsuperscript{\rm 1},
    Sachith Seneviratne\textsuperscript{\rm 1,\rm 2},
    Malin Premaratne\textsuperscript{\rm 3},
    Saman Halgamuge\textsuperscript{\rm 1}
}
\affiliations{
    \textsuperscript{\rm 1}AI, Optimization and Pattern Recognition Research Group, Dept. of Mechanical Eng., University of Melbourne, Australia \\
    \textsuperscript{\rm 2} Melbourne School of Design, University of Melbourne, Australia \\
    \textsuperscript{\rm 3}Department of Electrical and Computer Systems Engineering
, Monash University, Australia
    nsranasinghe@student.unimelb.edu.au, \{damith.senanayake, sachith.seneviratne\}@unimelb.edu.au, malin.premaratne@monash.edu, saman@unimelb.edu.au
}

\usepackage{bibentry}
\begin{document}

\maketitle

\begin{abstract}
Traditional machine learning is generally treated as a black-box optimization problem and does not typically produce interpretable functions that connect inputs and outputs. However, the ability to discover such interpretable functions is desirable. In this work, we propose GINN-LP, an interpretable neural network to discover the form and coefficients of the underlying equation of a dataset, when the equation is assumed to take the form of a multivariate Laurent Polynomial. This is facilitated by a new type of interpretable neural network block, named the “power-term approximator block”, consisting of logarithmic and exponential activation functions. GINN-LP is end-to-end differentiable, making it possible to use backpropagation for training. We propose a neural network growth strategy that will enable finding the suitable number of terms in the Laurent polynomial that represents the data, along with sparsity regularization to promote the discovery of concise equations. To the best of our knowledge, this is the first model that can discover arbitrary multivariate Laurent polynomial terms without any prior information on the order. Our approach is first evaluated on a subset of data used in SRBench, a benchmark for symbolic regression. We first show that GINN-LP outperforms the state-of-the-art symbolic regression methods on datasets generated using 48 real-world equations in the form of multivariate Laurent polynomials. Next, we propose an ensemble method that combines our method with a high-performing symbolic regression method, enabling us to discover non-Laurent polynomial equations. We achieve state-of-the-art results in equation discovery, showing an absolute improvement of 7.1\% over the best contender, by applying this ensemble method to 113 datasets within SRBench with known ground-truth equations.
\end{abstract}

\section{Introduction}
\label{intro}

Although supervised machine learning (ML) can create arbitrary mappings between a large number of input variables and target variables, typically, there is no ability to discover an interpretable mathematical expression between the two variable spaces. However, in cases where the recovery of an underlying governing equation is required, this lack of interpretability is an obstacle. Interpretable machine learning has the potential to harness data to discover underlying equations that govern them. 

In this work, we focus on the data-driven discovery of equations that take the form of multivariate Laurent polynomials (LP). LPs produce equations important in physics and real-world systems, with some examples shown in Table \ref{table:example_equations}. 

\begingroup
\renewcommand{\arraystretch}{1.25} % Default value: 1
\begin{table}[t]
    \centering
    \begin{tabular}{c|c}
    \textbf{Phenomenon} & \textbf{Equation} \\%& n(var) & order\\
    \hline
         Coulomb's Law & $E_f = \frac{1}{4\pi\epsilon}\frac{q_1q_2}{r^2}$ \\%& 3 & 0\\
         Kinetic Energy &  $K = \frac{m}{2}(u^2 + v^2 + w^2)$\\% & 3 & 2 \\
         Potential Energy (Gravity) & $U = Gm_1m_2(\frac{1}{r_2} - \frac{1}{r_1})$ \\%& 4 & 1\\
         \hline
    \end{tabular}
    \caption{Three example equations as multivariate LPs and their relevant phenomena}
    \label{table:example_equations}
\end{table}
\endgroup

More formally, we aim to discover a multivariate LP that can accurately map features $\textbf{X} \in \mathbb{R}^d$ to a target $y \in \mathbb{R}$ using a dataset of paired samples for $\textbf{X}$ and $y$. An exhaustive search is tractable to solve this problem when the search space is small but quickly becomes intractable as the number of variables and possible terms increases, as demonstrated in the appendix\footnote{Appendix available in https://arxiv.org/abs/2312.10913}. 

In this work, we present GINN-LP, a Growing Interpretable Neural Network for discovering multivariate Laurent Polynomials, which can efficiently discover the form and coefficients of a multivariate LP equation that describes a given dataset, without explicitly searching through the equation space.

% The SR problem has been proven to be NP-hard \cite{Virgolin2022SymbolicNP-hard}. To tackle this computational intractability, SR is typically formulated to have a constrained search space and a method to search through this constrained space efficiently. Moreover, most existing methods require information on the order of the equation. Our work is motivated by the drawbacks of existing methods and the need to develop a method to discover underlying equations of data without this pitfall.

\textbf{The main contributions} of this work are as follows.
\begin{itemize}
    \itemsep0em
    \item A new type of interpretable neural network (NN) block named the ``power-term approximator" (PTA).
    \item GINN-LP: An interpretable neural network architecture that uses multiple PTA blocks to discover multivariate LPs using observed data.
    \item A neural network growth strategy allowing the automatic discovery of the number of terms in the underlying polynomial while reducing overfitting and training time.
    \item A model training strategy that can effectively train the interpretable neural network and decide the best trade-off between accuracy and model simplicity.
\end{itemize}

\subsection{Preliminaries}

Here, we describe the preliminaries in formulating our rationale.

\textbf{\textit{Definition}:} A multivariate LP has the form
\begin{equation}
    P = \sum_{i = 1} ^ {n} c_i \prod_{j=1}^k V_j^{p_i(j)}
    % P = \sum_{-n < k_1 + k_2 + ... + k_m < n}C_{k_1k_2...k_m}\Pi_{i=1}^m x_i^{k_i}
\label{laurent_poly}
\end{equation}

where $c_i \in \mathbb{R}$ is the i-th term of the polynomial and $p_i(j) \in \mathbb{Z}$ is the power of the j-th variable in that term. The set of variables $V_j \in \{ V_1, ..., V_k\}$ is the finite set of variables considered in the LP. 

% \textcolor{red}{
% We evaluate our method on symbolic regression datasets with real-world ground-truth equations. A few examples are given below.
% }

% \begin{itemize}
%     \item $E_f = \frac{1}{4\pi\epsilon}*\frac{q_1q_2}{r^2}$
%     \item $K = \frac{m}{2}(u^2 + v^2 + w^2)$
%     \item $U = Gm_1m_2(\frac{1}{r_2} - \frac{1}{r_1})$
% \end{itemize}

To the best of our knowledge, GINN-LP is the first method which can discover governing equations with arbitrary multivariate LP terms. The NN is end-to-end differentiable and therefore, can be trained using backpropagation. A neural network growth strategy is implemented to automatically discover the correct number of terms in the underlying equation. The model is trained on a dataset consisting of $(\mathbf{x_i}, y_i)$ pairs (input-output) where $0 \leq i \leq N$ and N is the number of data points in the dataset. 

% We note that although we mainly focus on data with underlying multivariate LP equations, learning polynomials from data may be useful even when the underlying equation governing the dataset is not a polynomial. For instance, in approximation theory, polynomials are considered to be good approximators of more complex functions, including solutions to differential equations \cite{Funaro1992PolynomialEquations}. However, this work focuses mainly on data with underlying multivariate LPs equations.

% Our symbolic NN can learn,
% \begin{itemize}
%     \itemsep0em 
%     \item The structure of the equation (terms of the polynomial)
%     \item Coefficients of the equation
% \end{itemize}
% simultaneously when the underlying equation is a LP $f(\textbf{x})$ that maps the input to the output. This equation can be recovered using the weight matrix of the trained NN.

In the following section, we review the existing literature on related methods. Then, we introduce GINN-LP, our interpretable NN that can discover multivariate LPs describing datasets. The results of the experiments conducted are then presented, and the final section provides a summary of this work, its limitations and some future research directions.

\section{Related Work}

Data-driven discovery of equations has been mostly dominated by genetic programming (GP) methods in the early literature, exemplified by the fact that symbolic regression (SR) was first introduced as a GP problem in \cite{JohnR.Koza1992GeneticSelection}. However, with the increased interest in explainable AI, a variety of methods have been developed in the recent past \cite{LaCava2021ContemporaryPerformance}. 

\textbf{Methods for symbolic regression}: Symbolic regression is an area of research focused on discovering equations that describe given datasets, by efficiently searching through the space of possible equations. 

A well-known method named AIFeynman discovers simplifying properties (e.g. symmetry, separability) of real-world equations using a NN to greatly reduce the search space till the search space becomes tractable \cite{Udrescu2020AIRegression, Udrescu2020AIModularity}.  Another class of SR methods use sparse regression such as least absolute shrinkage and selection operator (LASSO) \cite{Tibshirani1996RegressionLasso} to recover a sparse coefficient matrix of a pre-defined equation structure, making the overall search more efficient \cite{Brunton2015DiscoveringSystems, Rudy2017Data-drivenEquations, Brunton2016SparseSINDYc, Kaheman2020SINDy-PI:SINDy-PI, Quade2018SparseRecovery, Messenger2021LearningWSINDy}. Sparse regression-based methods for equation recovery have been further improved using Bayesian methods \cite{Zhang2018RobustBars, Jin2019BayesianRegression}, allowing the quantification of uncertainty.

\textbf{Genetic programming (GP)}: Most early SR methods use genetic programming \cite{Koza1994GeneticSelection, Dubcakova2011Eureqa:Review, Schmidt2009DistillingData}. In these algorithms, a population of equations are evolved along multiple iterations till an acceptably accurate equation is found. GP based methods have been successfully used in areas such as microbial modelling \cite{Vidanaarachchi2020IMPARO:Optimisation}. GP algorithms suffer from the combinatorial nature of the problems, i.e., the search space is vast and a large number of equally suitable equations may be recovered. Additionally, GP algorithms tend to be slower in convergence compared to other optimization methods such as gradient-based methods and therefore do not scale well. 

\textbf{Shallow Neural Networks}: A class of equation learning neural networks, called equation learners (EQL) is presented in \cite{Martius2016ExtrapolationEquations, Sahoo2018LearningControl}. In EQL, the conventional activation functions are replaced by operators such as summation and trigonometric functions which are commonly found in real-world equations. This EQL network is integrated with deep learning architectures for end-to-end training in \cite{Kim2021IntegrationDiscovery}. PDE-NET 2.0 \cite{Long2018PDE-NetNetwork} uses a symbolic neural network inspired by \cite{Martius2016ExtrapolationEquations} to approximate the non-linear response function of the underlying partial differential equation (PDE) governing a system. 

\textbf{Deep Neural Networks}: Some recent works in SR circumvent the uninterpretable nature of deep neural networks (DNN) by using a large model to search the smaller space of symbolic models. Deep symbolic regression \cite{Petersen2021DeepGradients} proposes the use of a recurrent neural network (RNN) to output a distribution over symbolic expressions. The network is trained using a reinforcement learning method to learn the structure of the expression and the constants are learned using a non-linear optimization method such as the Broyden–Fletcher–Goldfarb–Shanno algorithm (BFGS) \cite{Fletcher2000PracticalOptimization}. 

In some recent deep learning methods, the model is pre-trained on a large collection of symbolic regression datasets with known equations  \cite{Valipour2021SymbolicGPT:Regression, Biggio2021NeuralScales, Kamienny2022End-to-endTransformers}. This contrasts with the majority of other methods, where the algorithms have to be trained from scratch for each SR dataset. Though they have been shown to be faster in equation discovery, the pre-training time could be quite large. % For example, the model introduced in \cite{Biggio2021NeuralScales} is trained for 3 days on a high-power GPU.

\textbf{Performance estimation and benchmarking}: The field of symbolic regression lacks a widely agreed-upon benchmarking platform or dataset. A recent work has introduced SRBench, an open-source, reproducible platform for benchmarking symbolic regression methods \cite{LaCava2021ContemporaryPerformance}, aiming to establish a standardized framework for evaluating SR methods. SRBench has incorporated over 130 datasets with ground truth equation forms to the Penn Machine Learning Benchmark (PMLB) \cite{Olson2017PMLB:Comparison}, which can be used to evaluate SR methods. The main source of symbolic regression datasets used in SRBench is the Feynman dataset, which originates from the Feynman lecture series on physics \cite{Feynman2011TheHeat}. SRBench compares 14 different symbolic regression algorithms using these SR datasets as well as black-box datasets. 

\section{Method}

\begin{figure}[t]
    \centering
    \includegraphics[width=1\columnwidth]{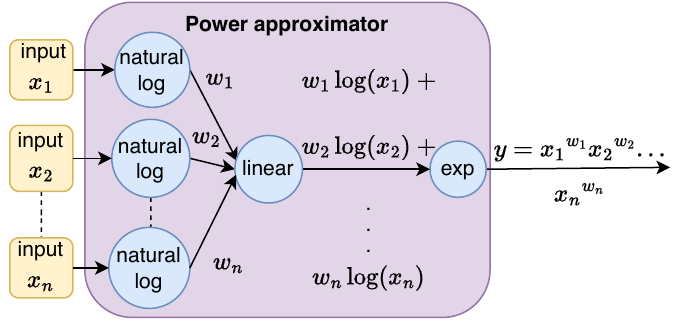}
    \caption{The architecture of the proposed interpretable NN block named the ``PTA" block. This block can discover a single term in a multivariate LP. $x_1, x_2,...,x_n$ are the inputs to the block and $w_1, w_2,...,w_n$ are weights of the linear activated neuron.}
    \label{fig:power_approximator}
\end{figure}

\begin{figure*}[t]
    \centering
    \includegraphics[width=1\textwidth]{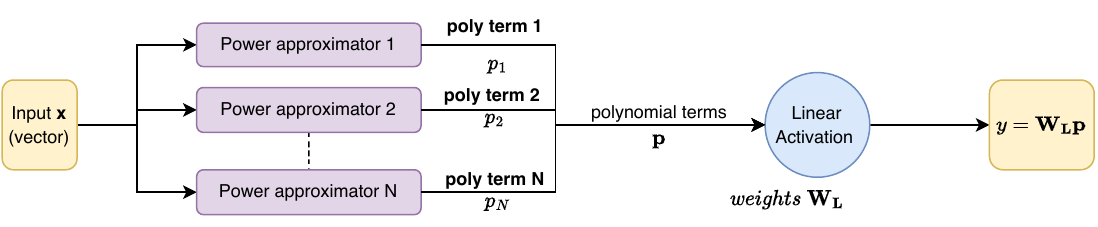}
    \caption{The architecture of the proposed interpretable neural network, GINN-LP. This consists of multiple PTA blocks in parallel, each discovering a single term in the underlying multivariate LP.}
    \label{fig:symbolic_nn}
\end{figure*}

\begin{algorithm}[tb]
\caption{GINN-LP training strategy}
\label{alg:algorithm}
\textbf{Input}: Dataset of input \textbf{x}, output y pairs\\
\textbf{Parameters}: Training instance count ($N=4$),  model complexity weight ($\alpha=10^{-6}$), $L_1$ regularization factor ($\lambda_1=10^{-4}$), $L_2$ regularization factor ($\lambda_2=10^{-4}$), maximum number of PTA blocks ($P=4$), number of epochs per network growth stage ($E$), rounding precision ($\epsilon)$\\
\textbf{Output}: An equation that describes the relationship between inputs and outputs
\begin{algorithmic}[1] %[1] enables line numbers
\FOR{$k_{inst} \gets 1$ to $N$}                    
    \STATE {model $\gets$ InitNetwork(num\_PTA=0)}
    \STATE {$MSE$ $\gets$ $\infty$}
    \STATE $max\_SE_M \gets \infty$
    \FOR{$k_{PTA} \gets 1$ to $P$}
        \STATE GrowModel(model)
        \STATE {TrainModel(model, $L_1$=$\lambda_1$, $L_2$=$\lambda_2$, epochs=$E/2$)}
        \STATE {TrainModel(model, $L_1$=$0$, $L_2$=$0$, epochs=$E/2$)}
        \STATE {RoundExponents(model, round\_prec=$\epsilon$)}
        \STATE $new\_MSE$ = CalcMSE(model, validation\_data)
        \IF {$new\_MSE > MSE*0.8$}
            \STATE {break}
        \ENDIF
        \STATE {$MSE = new\_MSE$}
    \ENDFOR
    \STATE $eq \gets $ GetEQ(model)
    \STATE $C_M \gets$ CalcComplexity(eq)
    \STATE $SE_M \gets MSE + \alpha * C_M$
    \IF {$SE_M > max\_SE_M$}
        \STATE {$max\_SE_M \gets SE_M$}
        \STATE {$best\_eq \gets eq$}
    \ENDIF
\ENDFOR
\STATE {\textbf{return} $best\_eq$}
% \ENDWHILE
% \STATE \textbf{return} solution
\end{algorithmic}
\end{algorithm}

In this work, we present GINN-LP, a novel growing interpretable neural network that can discover multivariate LP equations which describe a dataset\footnote{Source code available in https://github.com/nisalr/ginn-lp}. Once trained on a dataset of input-output pairs, the network can produce a LP that fits the data, while accurately predicting whether the LP assumption is valid for the given dataset.

The main component of the proposed architecture is the ``power-term approximator" (PTA) block, which is an interpretable NN block consisting of a logarithmic unit and a single linear activated neuron followed by an exponential activation. We illustrate the architecture of a PTA in Figure \ref{fig:power_approximator}. If $x_1, x_2, ..., x_n$ denote the inputs to the block, the output $y_d$ of the linear activated neuron and the output of the network $y$ are given by,

\begin{align}
    \label{yd_eq}
    y_d & = w_1\log(x_1) + w_2\log(x_2) + ... + w_n\log(x_n) \\
        & = \log(x_1^{w_1}x_2^{w_2}...x_n^{w_n})\\
    y &= e^{y_d} = x_1^{w_1}x_2^{w_2}...x_n^{w_n}
\end{align}

% \begin{align}
% \label{yd_eq}
%     y_d = w_1log(x_1) + w_2log(x_2) + ... + w_nlog(x_n) 
%     = log(x_1^{w_1}x_2^{w_2}...x_n^{w_n})
% \end{align}

where $w_1, w_2, ..., w_n$ are the weights of the linear activated neuron. 

% \begin{align}
% \label{y_eq}
    
% \end{align}

Since the PTA can be generalized to support any arbitrary number of inputs, a single block can in theory approximate an arbitrary multivariate LP with a single term. 

To enable the network to discover equations with multiple terms, a set of PTA blocks are stacked in parallel, followed by a single neuron with a linear activation as shown in Figure \ref{fig:symbolic_nn}. The output is a linear combination of the outputs of all PTA blocks. Each PTA is expected to discover a single term in a polynomial equation, including linear terms. More generally, they can recover equations with multiple additive terms each having an arbitrary number of variables raised to arbitrary powers. To promote the learning of exact equations, we perform a rounding of all coefficients and exponents to the nearest $\epsilon$ after training. $\epsilon$ was empirically set to 0.001 in our experiments.

\subsection{Training Strategies}

In our implementation, GINN-LP is trained to minimize the mean squared error (MSE) using the Adam optimizer and an exponentially decaying learning rate starting from 0.01.
Algorithm \ref{alg:algorithm} outlines the training strategies used to train GINN-LP efficiently, along with the empirically determined hyperparameters. We further discuss these strategies below.

\textbf{Regularization to enforce equation conciseness}.
In practice, simpler equations are preferred over more complex ones to describe relationships between variables since they are easier to understand and interpret. To enforce this property, we impose sparsity-promoting regularization on our interpretable neural network. Since the values of the weight matrix correspond to the coefficients and power terms of the discovered equation, a sparser weight matrix would result in a simpler equation. 

However, using regularization throughout training from beginning to end causes some discovered equations to be slightly different to the ground-truth equation. To avoid this, we introduce an unregularized training phase after regularized training. This allows the equation coefficients to converge towards more accurate values.

During the regularized training phase, a linear combination of $L_1$ and $L_2$ regularization is applied. Hence, the objective function would be,

\begin{align}
\label{loss_eq}
    L = \frac{1}{N}\sum_{i=1}^{N}(f(\textbf{x}) - y_i)^2 + \lambda_1\sum_{i=1}^n|W_i| + \lambda_2\sum_{i=1}^nW_i^2
\end{align}

Here, N is the number of samples in the dataset, n is the number of weights in the weight matrix and $\lambda_1$, $\lambda_2$ are the regularization constants. When the unregularized phase was not included, we observed that the equation coefficients would sometimes deviate from their true values by small amounts (e.g. 0.001). The point at which the regularization changes is expressed as a fraction of the total number of training epochs. This was empirically set to 0.5.

\textbf{Neural network growth}. Having a higher number of PTA blocks than required could lead to the model overfitting. Therefore, starting from a single PTA block, the number of PTA blocks is increased, thereby growing the GINN-LP architecture. When the network is grown, a new, randomly initialized PTA block is added in parallel without altering the weights of the already trained PTA blocks. Then, the grown network is trained for a fixed number of epochs on the dataset. This is performed iteratively till the network is grown to a pre-defined maximum size, or till an early stopping condition is reached.

% An early stopping condition for the neural network growth is defined to stop the training at the correct number of PTA blocks required to recover the exact governing equation. This is expected to be equal to the number of terms in the polynomial that describes the dataset. 
Training is stopped if the validation MSE does not decrease by a certain percentage after each growth iteration. This percentage is a hyperparameter of the model and is empirically set to 20\% in our experiments. We expect the training to stop when the number of PTA blocks is equal to the number of terms in the equation.

\textbf{Training multiple instances}. When training GINN-LP, the network sometimes gets stuck in local optimums. To ensure a well-fitting model, we train the network multiple times with different random initializations, and the best model is selected as outlined in the following sub-section.

\textbf{Model selection}. When there are multiple models (equations) with similar performances on unseen data, we prefer the simpler model. To select the best model from all training instances, we propose a new performance metric named the symbolic error (SE) using a linear combination of the model complexity, and the validation set MSE. We use the same complexity metric as SRBench \cite{LaCava2021ContemporaryPerformance}, where the complexity ($C_M$) of a given model $M$ is defined simply as the number of mathematical operators ($\mathrm{op}_M$), constants ($\mathrm{const}_M$) and features ($n_m$) in the equation discovered by the model. The best model $\hat{M}$ is determined as shown below,

\begin{align}
\label{complexity_eq}
    C_M &= \mathrm{op}_M + \mathrm{const}_M  + n_M \\
    SE_M &= MSE(y, \hat{y} | \theta_M) + \alpha * C_M \\
    \hat{M} &= \argmin_M[SE_M]
\end{align}

Here, $\theta_M$ are the parameters of model $M$, $y$ is the target, $\hat{y}$ is the predicted value and $\alpha$ is the weighting factor of the model complexity. $\alpha = 10^{-6}$ is determined empirically.

\subsection{Ensemble Method for Discovering Non-Laurent Polynomial Equations}

\begin{figure}[t]
    \centering
    \includegraphics[width=1\columnwidth]{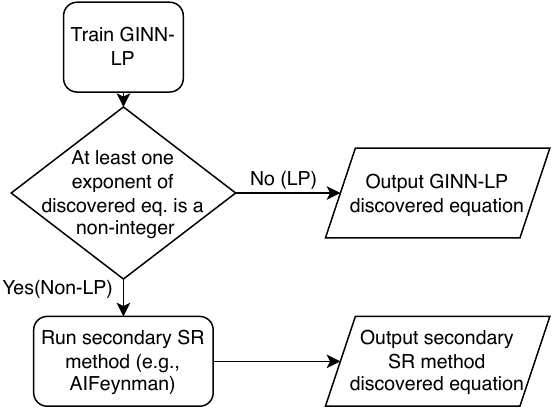}
    \caption{ The ensemble pipeline combines GINN-LP with another high-performing SR method, enabling it to discover both LP and non-LP equations}
    \label{fig:ensemble_pipeline}
\end{figure}

To demonstrate the broader applicability of GINN-LP, we propose an ensemble equation discovery pipeline that combines GINN-LP with high-performing SR methods, to enable the discovery of both LP and non-LP equations, thereby achieving state-of-the-art performance in equation discovery. In this ensemble method, we first fit GINN-LP, and if at least one of the exponents of the discovered equation is not an integer, we determine that the assumption of the ground-truth equation being an LP is invalid. In this case, we fit a secondary SR method to discover an accurate equation. The ensemble pipeline is illustrated in Figure \ref{fig:ensemble_pipeline} and two hypothetical examples showing the GINN-LP and secondary SR method outputs are given in Table \ref{tab:ensemble_examples}.

We perform multiple experiments with different options for the secondary SR method, and show that ensembling with GINN-LP improves the performance of all other SR methods used for evaluation.

\begin{table}[t]
    \centering
    \begin{tabular}{|p{1.8cm}|p{1.5cm}|p{1.5cm}|p{1.5cm}|}\hline
 GINN-LP Output& LP/Non-LP? & Secondary SR output&Ensemble output\\\hline 
         $x_1^{2.342}x_2^{2}$&  Non-LP&  $sin(x_1)/x_2$ & $sin(x_1)/x_2$\\ \hline 
         $x_1^2x_2^{-1}$&  LP&  N/A & $x_1^2x_2^{-1}$\\ \hline
    \end{tabular}
    \caption{Hypothetical examples of the ensemble method output, showing the intermediate outputs.}
    \label{tab:ensemble_examples}
\end{table}

\section{Results}

We evaluate our approach on the Feynman symbolic regression benchmark dataset, using SRBench \cite{LaCava2021ContemporaryPerformance}. The Feynman dataset consists of multiple SR datasets generated from real-world equations, along with their corresponding ground-truth equations. We note that our method may not qualify as a symbolic regression method, since we are not explicitly searching through a space of equations. However, due to the similarities in the aim, we evaluate our method against SR methods. 

GINN-LP was compared against 14 other popular SR methods in terms of how well it recovers the ground-truth equation. Each SR method was run for five trials per equation dataset to investigate whether they can consistently recover the correct ground truth equation with different random initializations.

All experiments were conducted on a computing cluster where each experiment used a 32-core, 2.90GHz Intel Xeon Gold 6326 CPU and a single NVIDIA A-100 GPU. We note that our approach does not see a major performance improvement when training on a GPU, since the network architecture is small. Moreover, we did not re-run experiments for other methods where results were already available within SRBench \footnote{SRBench code available in https://github.com/cavalab/srbench}.

\subsection{Performance Metric}

We use the same definition of exact recovery as SRBench \cite{LaCava2021ContemporaryPerformance}. Each recovered equation is simplified using SymPy \cite{Meurer2017SymPy:Python} and compared with the corresponding ground-truth equation to determine whether it has been recovered exactly. For each SR method tested, the performance is reported using the symbolic solution recovered percentage, also called the solution rate. This is calculated as,
\begin{align}
\label{best_model}
    solution\ rate = \frac{correctly\ recovered\ equations}{total\ number\ of\ equations} \times 100\%
\end{align}
Since multiple trials (with different random initializations) are performed for each dataset and algorithm, we report a range of the solution rate or the mean/median value. In each trial, the network was trained on 10,000 data points, generated from the ground-truth equation of each of the datasets.

\begin{figure}[t]
    \centering
    \includegraphics[width=1\columnwidth]{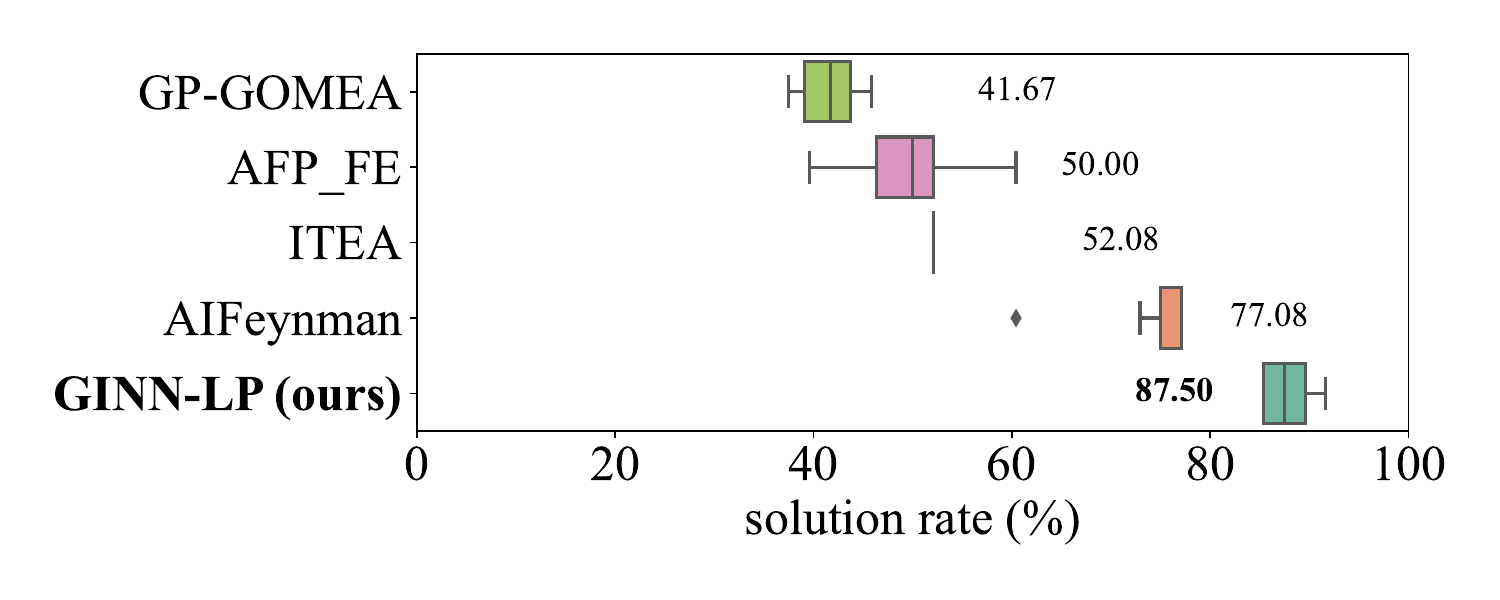}
    \caption{Solution rate of the top five algorithms, for all datasets with LP ground-truths. The median solution rates are shown on the side of each plot.}
    \label{fig:symnn_LP}
\end{figure}

\begin{figure}[t]
    \centering
    \includegraphics[width=1\columnwidth]{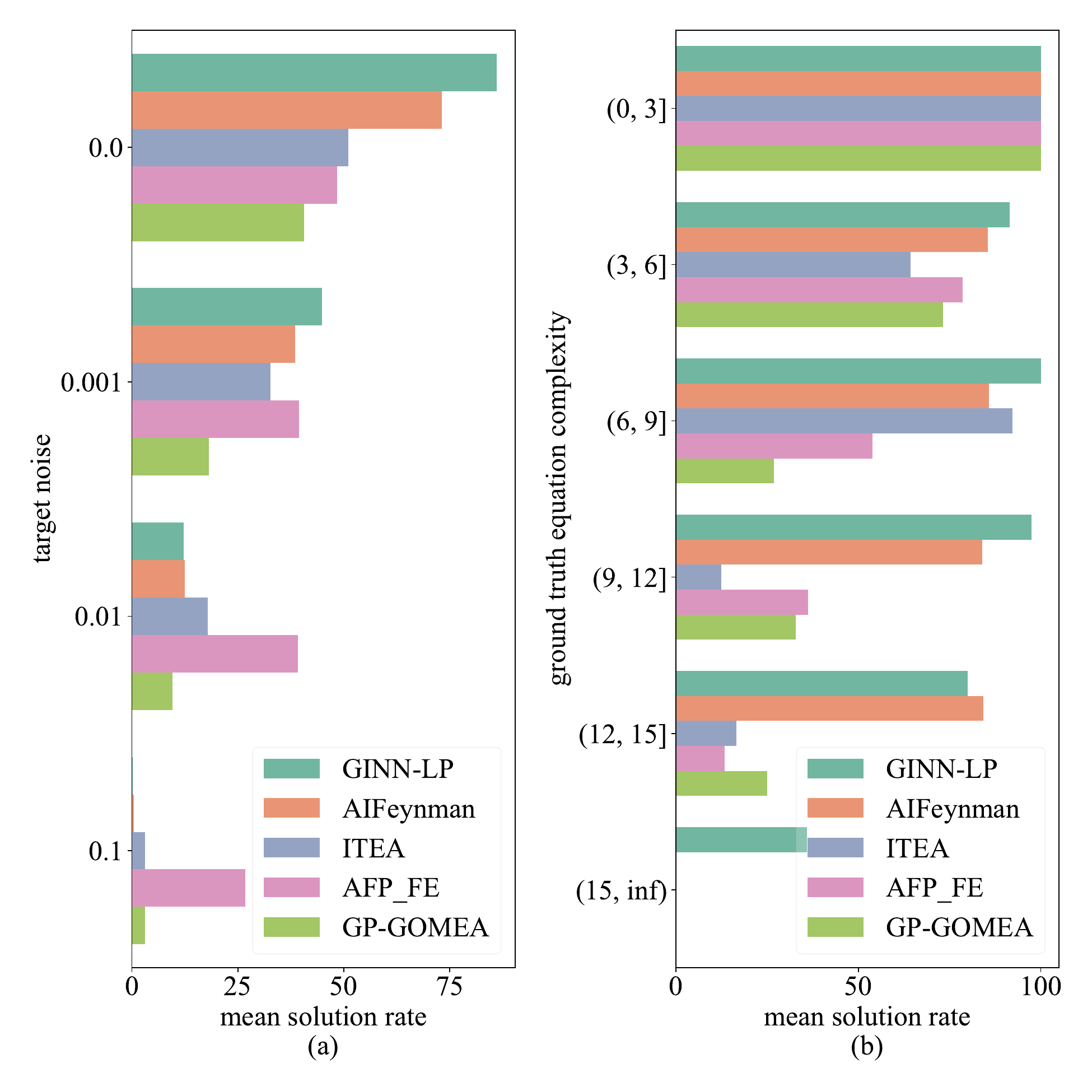}
    \caption{ (a) Comparison of SR methods, with the presence of target noise. The mean solution rates are reported. (b) Comparison of SR methods with respect to the ground truth equation complexity. The mean solution rate is reported}
    \label{fig:results_noise_comp}
\end{figure}

\begin{figure*}[t]
    \centering
    \includegraphics[width=1\textwidth]{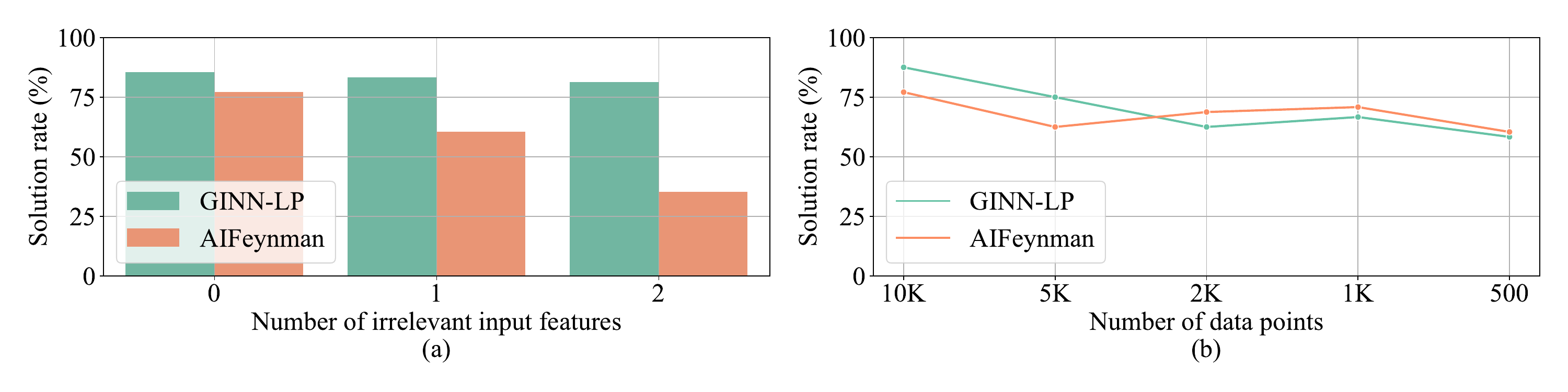}
    \caption{(a) Model performance (solution rate) with respect to the number of irrelevant inputs features (b) Model performance (solution rate) with a varying number of data points used for training}
    \label{fig:results_red_in_samp}
\end{figure*}

\subsection{Standalone GINN-LP Performance Results – Recovering LP Equations}

We first selected the subset of 48 multivariate LPs within the Feynman dataset to evaluate our method, as GINN-LP is designed to retrieve such equations. The ground-truth equations of these datasets are listed in the appendix. 

\textbf{Comparison of models without added noise}. We compare GINN-LP with 14 popular symbolic regression algorithms using SRBench. The results of the experiments without adding any noise to the datasets are shown in Figure \ref{fig:symnn_LP} for the five top-performing algorithms. Our method outperforms all the 14 algorithms it was compared against in terms of minimum, maximum and median solution rates. To ensure fairer comparison, experiments were performed by reducing the search space of AIFeynman only to include operators present in LPs, to incorporate the same prior that we are incorporating into our method. However, the performance of AIFeynman did not increase noticeably with this reduced search space.

\textbf{Impact of adding Gaussian noise}. To evaluate the noise-resilience of GINN-LP, experiments were conducted after adding Gaussian white noise to the target as a fraction of the target root mean square value. Experiments were repeated with noise fractions of 0.001, 0.01 and 0.1, on the top five algorithms identified in the previous section (without adding noise) and the results are shown in Figure \ref{fig:results_noise_comp}(a). Our method outperforms the other top methods with low noise (0.001) but does not manage to beat the AFP\_FE method with high noise values. 

Equations that were recovered by our method in the high noise cases, seemed to have recovered the structure with some slight changes to the constants. We hypothesize that the network overfits due to attempting to model the noise.

However, we note GINN-LP is still able to model noisy data quite accurately, even if it’s unable to recover the exact groundtruth equations, as shown in the appendix.

\textbf{Solution rate vs ground truth equation complexity}. Further experiments were conducted to observe how the GINN-LP performance changes with the complexity of the equation we attempt to recover. Here, complexity $C_M$ is calculated using the metric defined in Equation \ref{complexity_eq}. The equations in our dataset consisted of complexity values ranging from 3 to 27. To visualize our results, we grouped the complexity values into six bins. The intervals of these bins are $(0, 3], (3, 6], (6, 9], (9, 12], (12, 15]$ and $(15, +\infty)$. 

We compare the performance of the five top-performing algorithms on each of the six bins. The mean solution rate (across all experiments) for each bin for a particular algorithm is visualized in Figure \ref{fig:results_noise_comp}(b). Our method maintains good solution rates across most of the complexity bins, with a moderate solution rate of close to 40\% even for the most complex bin, where all other methods fail. 

\textbf{Training with irrelevant inputs}. In practice, we might not know which of our input variables are related to the output. We investigate how GINN-LP performs when there are such variables, by adding irrelevant inputs sampled from a uniform distribution within the same range as the other inputs. Experiments were conducted with up to 2 irrelevant inputs and the results are shown in Figure \ref{fig:results_red_in_samp}(a). We compare our method with AIFeynman, which was the top-performing method (among other methods) in earlier experiments. Our method does not see a significant drop in performance even with 2 irrelevant inputs, while the performance of AIFeynman drops considerably. 

\textbf{Effect of training data size}. We compared our method with AIFeynman when trained with different sizes of training datasets and present the results in Figure \ref{fig:results_red_in_samp}(b). We note that GINN-LP can still recover a majority of the equations even with a small number of samples, and remains competitive with AIFeynman.

\textbf{Effect of epoch count and number of training instances}. Further experiments were done to investigate the effect of these hyperparameters on the performance of GINN-LP. Results of these experiments are provided in the appendix.

\subsection{Ensemble Method Performance Results}

We evaluate the performance of the proposed ensemble method using the entire Feynman symbolic regression benchmark dataset within SRBench. This consists of 113 datasets, including all 48 datasets with LP ground-truths. GINN-LP perfectly identifies when the LP assumption does not hold for a given dataset, as demonstrated by the confusion matrix included in the appendix. This indicates that GINN-LP can successfully identify the problems it cannot solve, enabling them to be solved using a secondary, high-performing SR method. We show the results after ensembling our method with AIFeynman and Multiple Regression Genetic Programming (MRGP) in the below sub-sections, since these two methods show the best performances in terms of their solution rate and predictive accuracy respectively. We further show in the appendix, that ensembling with GINN-LP enhances the performance of all 14 other SR methods.

\begin{figure}[t]
    \centering
    \includegraphics[width=1\columnwidth]{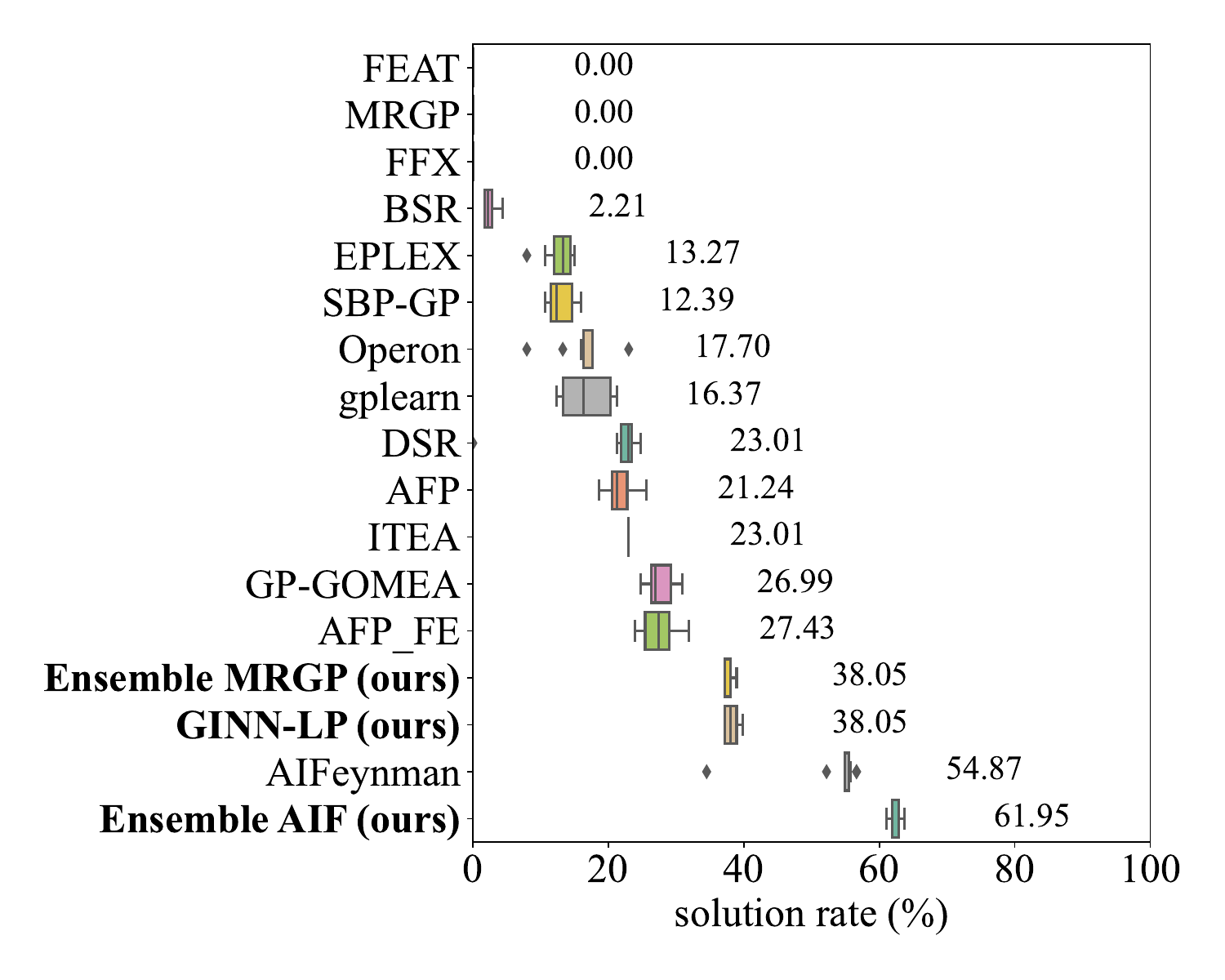}
    \caption{Solution rate of ensemble models (with MRGP, AIFeynman) and standalone GINN-LP, compared against other methods, for all 113 datasets in the Feynman symbolic regression benchmark dataset (including non-LP). The median solution rate is shown on the side of each plot.}
    \label{fig:hybrid_method}
\end{figure}

\textbf{Solution rate}. The results of the proposed ensemble methods in terms of their solution rates are shown in Figure \ref{fig:hybrid_method}. We show that our method ensembled with AIFeynman outperforms state-of-the-art methods by 7.1\% in terms of the median solution rate (absolute). Moreover, despite being formulated for recovering LP equations, the standalone GINN-LP still outperforms the majority of other methods on the complete Feynman dataset.

\begin{figure}[t]
    \centering
    \includegraphics[width=1\columnwidth]{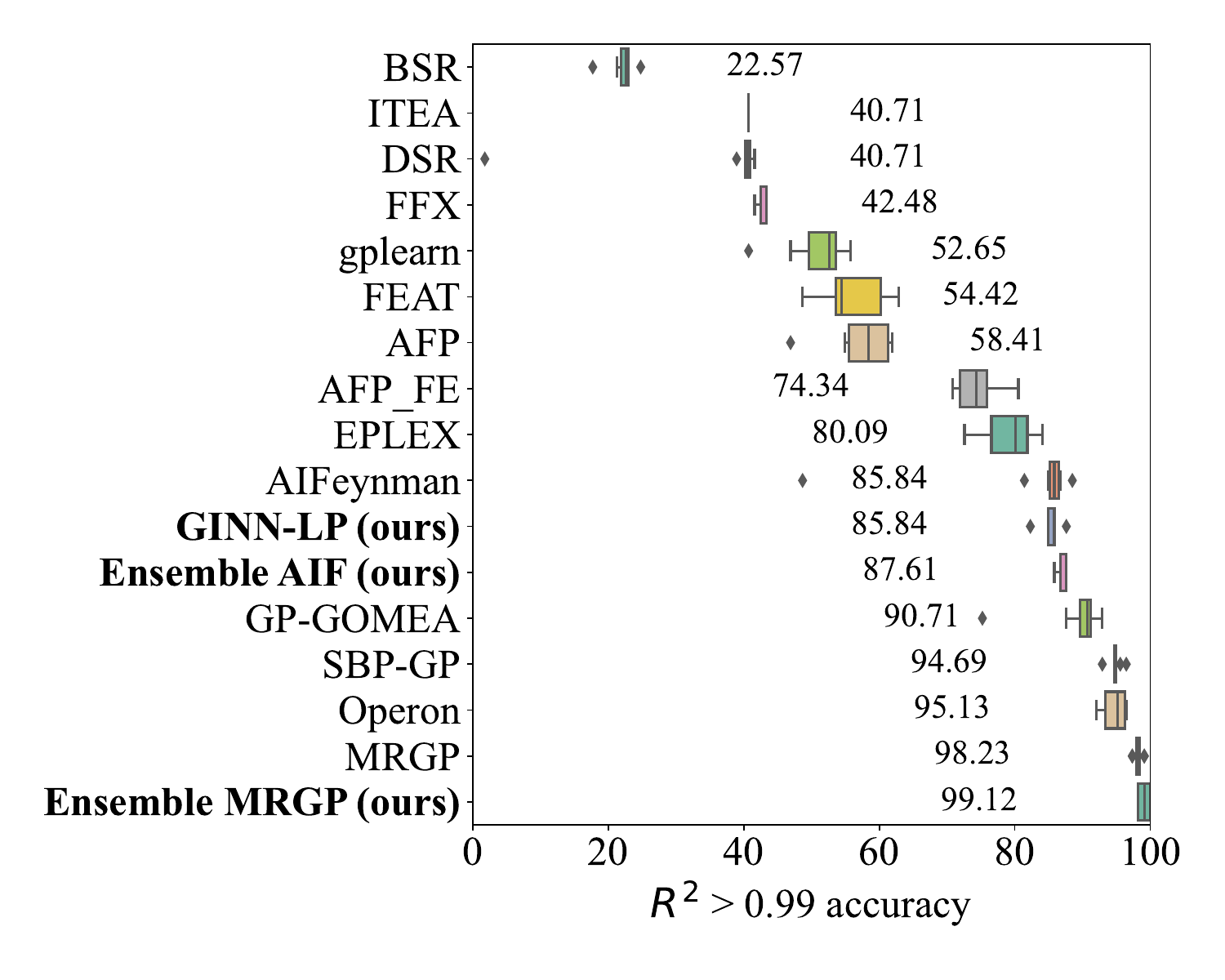}
    \caption{$R^2 > 0.99$ accuracy of ensemble models (with MRGP, AIFeynman) and standalone GINN-LP, compared against other methods, for all 113 datasets (including non-LP) in the Feynman symbolic regression benchmark dataset. The median accuracy is shown on the side of each plot.}
    \label{fig:hybrid_method_r2}
\end{figure}

\textbf{Predictive accuracy}. Further experiments were conducted to evaluate how accurately our methods can predict outputs for new, previously unseen data. The prediction accuracy is calculated as the percentage of datasets with $R^2$ above 0.99 on the test data. A train-test split of 75-25 is used. We compare the prediction accuracy of all methods on all 113 datasets, including GINN-LP, and the proposed ensemble methods. As shown in Figure \ref{fig:hybrid_method_r2}, all of our methods, including the standalone GINN-LP perform reasonably well while GINN-LP ensembled with MRGP outperforms all other methods by 0.9\%, while achieving near perfect $R_2 > 0.99$ accuracy.

\section{Conclusion}

In this work, we present GINN-LP, an end-to-end differentiable interpretable neural network that can recover mathematical expressions that take the form of multivariate LPs. This is made possible by a new type of interpretable neural network block we introduce, named the PTA block, that can discover terms of multivariate LPs. Our method can automatically discover the number of terms in the underlying equation using a neural network growth strategy while promoting the discovery of simpler equations by applying sparsity regularization.

The results of the experiments conducted show that our approach achieves the state-of-the-art results, outperforming the best contender in symbolic regression algorithms on datasets with multivariate LP ground-truth equations in the Feynman symbolic regression benchmark dataset. We also show that GINN-LP can successfully identify when a given dataset cannot be accurately described using a multivariate LP. Building on these observations, we propose an ensemble method which uses a secondary SR method to discover equations when GINN-LP identifies that the LP assumption does not hold for a given dataset. Further experiments show that this ensemble method achieves state-of-the-art results in equation discovery, when evaluated on the entire Feynman dataset, including both LP and non-LP equations.

A limitation of the proposed architecture is that it assumes that all input values inside and outside of the training dataset are positive real numbers. This is because of the natural logarithmic activation used within the PTA block. This could be overcome by adding a constant to all input features with negative values. However, this will make it more difficult to discover the mathematical equation since adding a constant to the input could change an LP equation to a more complex equation (e.g. $x_1/x_2$ changes to $(x_1 + 2)/(x_2 + 2)$). Future work may investigate NN architectures that can perform power approximation in the presence of negative input values. 

% This work could potentially be adapted to solving unsupervised learning problems. One approach is to train GINN-LP using unlabelled inputs and a fixed pre-defined output (e.g., 0), expecting it to recover an equation that relates the inputs such as $x^2 + y^2 + z^2 - 1 = 0$. However, this would likely produce a trivial result (e.g., 0*xy = 0) unless extra regularization is applied. Future work may look at extending this work to solve unsupervised learning problems.

Another interesting future research direction is to extend this architecture to discover a wider variety of equations, without being limited to only LPs. One way to achieve this is to grow the network sequentially, thereby stacking PTA blocks in multiple layers one after the other. However, this would significantly increase the risk of overfitting. Therefore, greater attention will need to be paid to regularizing the weights, in addition to the growth strategy.

\section{Acknowledgements}

This study was supported by the Melbourne Graduate Research Scholarship for the first author, the Australian Research Council grant DP210101135 for the first author and the Australian Research Council grant DP220101035 for the second, third and the last authors. This study was supported partially for the third author by the Australian Government through the Australian Research Council’s Centre of Excellence for Children and Families over the Life Course (Project ID CE200100025). This research was also supported by The University of Melbourne’s Research Computing Services and the Petascale Campus Initiative. The authors thank Nadarasar Bahavan, Deshani Poddenige and Rashindrie Perera for proofreading. 

\bibliography{references}

\newpage
\appendix
\onecolumn
\large{\textbf{Appendix}}
\renewcommand{\thesection}{\alph{section}}
\section{Exponential search space of polynomial equations}
\label{poly_search}

We demonstrate the intractable search space of multivariate polynomial structure by calculating the number of terms and possible structures for polynomials of a given order (n) and a given number of variables (k).

The number of terms $T$ can be calculated by,

\begin{equation}
    T = \binom{n+k}{n}
\end{equation}

The number of unique structures (S) of a polynomial with T terms is given by the number of unique subsets of these T terms. S is given by,

\begin{equation}
    S = 2^T
\end{equation}

We illustrate this in Figure \ref{fig:term_count} below.

\begin{figure}[h]
    \centering
    \includegraphics[width=1\columnwidth]{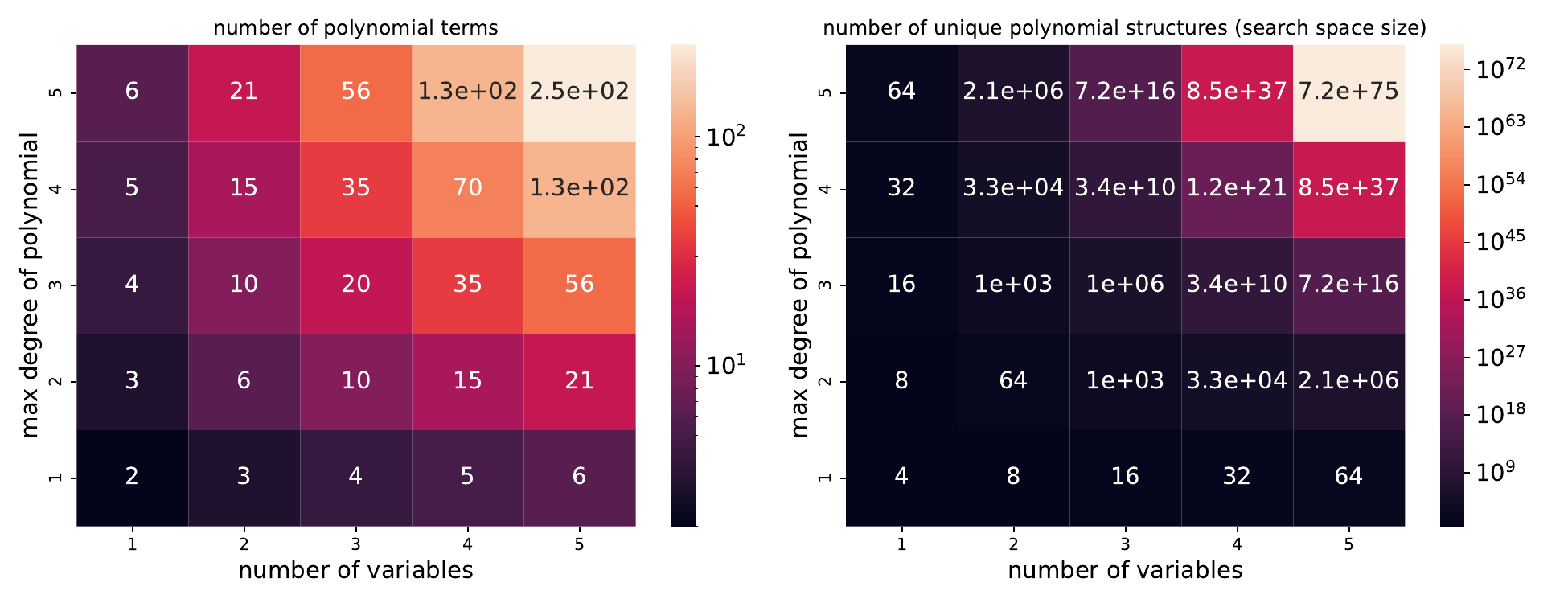}
    \caption{Number of terms in a polynomial (top) and the search space size of all unique polynomials (bottom) vs maximum degree and number of variables}
    \label{fig:term_count}
\end{figure}

\section{Identifying whether the ground-truth can be described using a Laurent polynomial}

\begin{figure}[h]
    \centering
    \includegraphics[width=0.4\columnwidth]{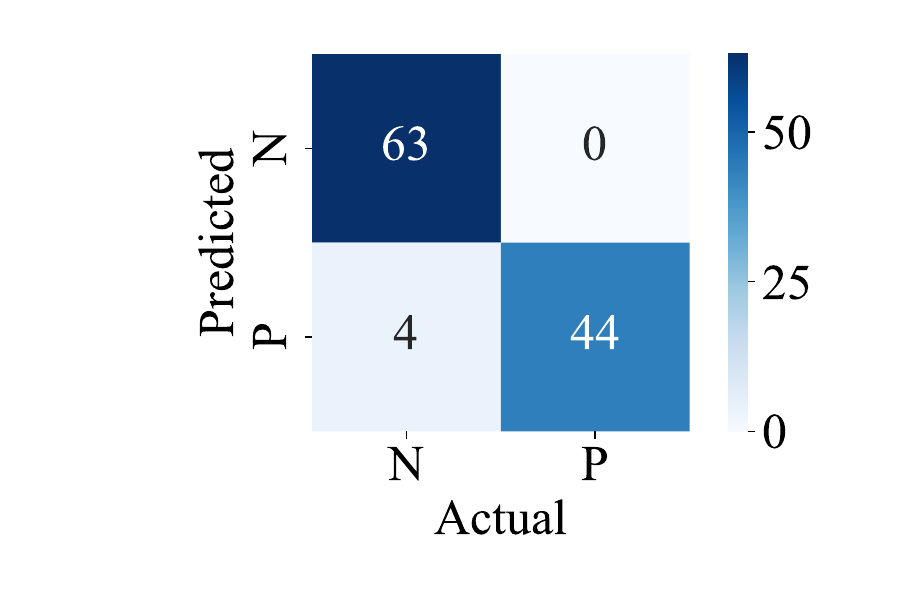}
    \caption{Confusion matrix of LP/Non-LP classification result. We use GINN-LP to identify whether the dataset can be described using a Laurent polynomial (LP)}
    \label{fig:conf_matrix}
\end{figure}

GINN-LP can accurately identify whether a given dataset can be described accurately using a LP. We train GINN-LP on all 113 datasets of the Feynman symbolic regression benchmark, within SRBench and classify them as LP or Non-LP. This is done by analyzing the exponents of the discovered equation. If all exponents are integers, we classify an equation as a LP. We compare the GINN-LP classification with the ground-truth equations and plot the confusion matrix in Figure \ref{fig:conf_matrix}. It was observed that there were zero False positives, demonstrating that when the LP assumption is invalid (the ground-truth is non-LP), GINN-LP can perfectly identify it.

\section{Laurent polynomial equations used for evaluating the model}
\label{eq_list}

The equations within SRBench \cite{LaCava2021ContemporaryPerformance} used to evaluate our method were taken from the Feynman symbolic regression benchmark dataset, which originates from the Feynman lecture series \cite{Feynman2011TheHeat}. Since our network is designed to discover multivariate LPs, we used the subset of LP equations from the Feynman dataset.

The 48 equations used to evaluate our method are given below.

\begin{enumerate}
    \item $E_f = \frac{1}{4\pi}*\frac{q_1q_2}{\epsilon r^2}$
    \item $U = gmz$
    \item $\frac{3}{8\pi G}(H_G^2 + \frac{c^2k_f}{r^2})$
    \item $P = \frac{A\kappa(-T_1 + T_2)}{d}$
    \item $K = \frac{m}{2}(u^2 + v^2 + w^2)$
    \item $A = x_1y_1 + x_2y_2 + x_3y_3$
    \item $E_{den} = E_f^2\epsilon$
    \item $m = \frac{({\frac{h}{2\pi}})^2}{E_nd^2}$
    \item $F = \frac{AYX}{d}$
    \item $U = Gm_1m_2(\frac{1}{r_2} - \frac{1}{r_1})$
    \item $\mu_m = \frac{qh}{4\pi m}$
    \item $F = \mu N_n$
    \item $F = \frac{q_1q_2}{4\pi\epsilon r^2}$
    \item $F = q_2E_f$
    \item $U = \frac{1}{2}k_{spring}x^2$
    \item $E_n = \frac{1}{2}(\omega^2 + \omega_0^2)x^2$
    \item $V_e = \frac{q}{c}$
    \item $k = \frac{\omega}{c}$
    \item $P = \frac{q^2a^2}{6\pi\epsilon c^3}$
    \item $E_n = \frac{h}{2\pi}\omega$
    \item $\omega = \frac{qvB}{p}$
    \item $r \frac{4\pi\epsilon h^2}{mq^2}$
    \item $\frac{3}{2}p_rV$
    \item $P_F = \frac{nk_bT}{V}$
    \item $v = \frac{\mu_{drift}qV_e}{d}$
    \item $D = \mu_ek_bT$
    \item $P_* = \frac{n_{\rho}p_d^2E_f}{3k_bT}$
    \item $B = \frac{1}{4\pi{\epsilon}c^2}\frac{2l}{r}$
    \item $F_e = \epsilon cE_f^2$
    \item $F_E = \frac{P}{4\pi r^2}$
    \item $\omega = \frac{g_{\_}qB}{2m}$
    \item $E = \frac{g_{\_}\mu_MBJ_z}{h}$
    \item $I = \frac{qv}{2\pi r}$
    \item $\mu_M = \frac{qvr}{2}$
    \item $f = \frac{\mu_mB}{k_bT} + \frac{\mu_m\alpha}{\epsilon c^2k_bT}$
    \item $E = \mu_M(1 + \chi)B$
    \item $V_e = \frac{q}{4\pi\epsilon r}$
    \item $E_{den} = \frac{\epsilon E_f^2}{2}$
    \item $E = \frac{3}{5}\frac{q^2}{4\pi\epsilon d}$
    \item $L = \frac{nh}{2\pi}$
    \item $v = \frac{4\pi E_nd^2k}{h}$
    \item $k = \frac{2\pi\alpha}{nd}$
    \item $E = \frac{-mq^4}{2(4\pi\epsilon)^2h^2}\frac{1}{n^2}$
    \item $j = \frac{-\rho_{c_0}qA_{vec}}{m}$
    \item $\omega = \frac{4\pi\mu_MB}{h}$
    \item $E_n = \frac{1}{2m}(p^2 + m^2\omega^2x^2(1+\frac{\alpha x}{y}))$
    \item $Pr = \frac{-1}{8\pi G}(\frac{c^4k_f}{r^2} + H_G^2c^2(1 - 2\alpha))$
    \item $P = -\frac{32}{5}\frac{G^4}{c^5}\frac{(m_1m_2)^2(m_1+m_2)}{r^5}$
    
\end{enumerate}

\section{Contemporary symbolic regression algorithms compared with the proposed interpretable neural network (GINN-LP)}
\label{algo_list}

Using SRBench, we compare our method with 14 symbolic regression algorithms seen in contemporary literature. These methods are listed below in Table \ref{algo_table}.

\begin{table}[h]
    \centering
    \caption{List of algorithms compared against GINN-LP}
    \begin{tabular}{|p{6cm}|p{4.5cm}|p{0.75cm}|p{4.25cm}|}
    \hline
        \textbf{Description} & \textbf{Method name} & \textbf{Year} & \textbf{Method family} \\ \hline
        Age fitness pareto optimization & AFP \cite{Schmidt2010Age-fitnessOptimization} & 2011 & Genetic programming (GP) \\ \hline
        Age fitness pareto optimization with fitness estimates & AFP-FE \cite{Schmidt2009DistillingData} & 2011 & Genetic programming (GP) \\ \hline
        AIFeynman & AIFeynman \cite{Udrescu2020AIRegression} & 2020 & Divide and conquer /  Neural networks \\ \hline
        Bayesian symbolic regression & BSR \cite{Jin2019BayesianRegression} & 2020 & Bayesian methods / Markov chain Monte Carlo \\ \hline
        Deep symbolic regression & DSR \cite{Petersen2021DeepGradients} & 2021 & Deep learning \\ \hline
        Epsilon lexicase selection & EPLEX \cite{LaCava2019ASelection}& 2016 & Genetic programming (GP) \\ \hline
        Feature engineering automation tool & FEAT \cite{LaCava2019LearningTrees} & 2019 & Genetic programming (GP) \\ \hline
        Fast function extraction & FFX \cite{McConaghy2011FFX:Technology}& 2011 & Random search \\ \hline
        GP version of the gene-pool optimal mixing evolutionary algorithm & GP-GOMEA \cite{Virgolin2021ImprovingExpressions} & 2020 & Genetic programming (GP) \\ \hline
        GP for symbolic regression in Python & gplearn & 2015 & Genetic programming (GP) \\ \hline
        Interaction transformation evolutionary algorithm & ITEA \cite{deFranca2021InteractiontransformationRegression}& 2020 & Genetic programming (GP) \\ \hline
        Multiple regression genetic programming & MRGP \cite{Arnaldo2014MultipleProgramming}& 2014 & Genetic programming (GP) \\ \hline
        SR with non-linear least squares & Operon \cite{Kommenda2020ParameterSquares}& 2019 & Genetic programming (GP) \\ \hline
        Semantic back-propagation genetic programming & SBP-GP \cite{Virgolin2019LinearRegression}& 2019 & Genetic programming (GP) \\ \hline
    \end{tabular}
    \label{algo_table}

\end{table}

\section{Comparison of SR methods in terms of $R^2 > 0.99$ accuracy, when a high amount of noise if present}

\begin{figure}[h]
    \centering
    \includegraphics[width=1\columnwidth]{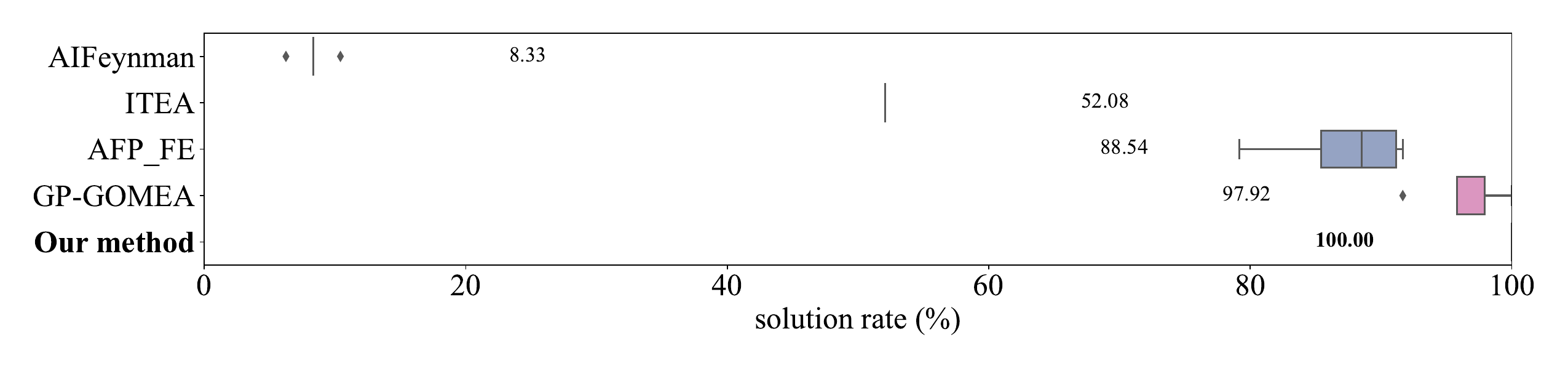}
    \caption{Comparison of $R^2 > 0.99$ accuracy of SR methods, under the highest noise ratio of 0.1}
    \label{fig:high_noise_r2}
\end{figure}

When a high amount of noise is present, GINN-LP solution rate drops as shown earlier in the paper. However, GINN-LP still performs very well in modelling the data, even though it may not recover the exact groundtruth equation. Further analysis of the model outputs show that GINN-LP achieves a $R_2 > 0.99$ accuracy of 100\%, even with the highest noise ratio of 0.1, as shown in Figure \ref{fig:high_noise_r2}.

\section{Performance comparison of GINN-LP, ensembled with all other methods}

We perform experiments after ensembling GINN-LP with all 14 contemporary methods used for comparison. Performance was evaluated using solution rate and $R_2 > 0.99$ accuracy as defined earlier. Almost all methods improve upon ensembling with GINN-LP as shown in Table \ref{ensemble_perf}.

\begin{table}[h]
    \centering
    \caption{Performance of SR methods after ensembling with GINN-LP}
    \label{ensemble_perf}
    \begin{tabular}{|p{2.75cm}|p{1.75cm}|p{1.75cm}|p{1.75cm}|p{1.75cm}|p{1.75cm}|p{1.75cm}|}
    \hline
        \textbf{SR Method} & \textbf{solution rate - standalone} & \textbf{solution rate - ensembled with GINN-LP} & \textbf{solution rate improvement} & \textbf{$R_2 > 0.99$ accuracy - standalone} & \textbf{$R_2 > 0.99$ accuracy - Ensembled with GINN-LP} & \textbf{$R_2 > 0.99$ accuracy improvement} \\ \hline
        AFP & 21.2\% & 42.5\% & 21.2\% & 58.4\% & 67.3\% & 8.8\% \\ \hline
        AFP\_FE & 27.4\% & 45.1\% & 17.7\% & 74.3\% & 78.8\% & 4.4\% \\ \hline
        AIFeynman & 54.9\% & 61.9\% & 7.1\% & 85.8\% & 87.6\% & 1.8\% \\ \hline
        BSR & 2.2\% & 38.1\% & 35.8\% & 22.6\% & 49.6\% & 27.0\% \\ \hline
        DSR & 23.0\% & 46.9\% & 23.9\% & 40.7\% & 57.5\% & 16.8\% \\ \hline
        EPLEX & 13.3\% & 40.7\% & 27.4\% & 80.1\% & 81.4\% & 1.3\% \\ \hline
        FEAT & 0.0\% & 38.1\% & 38.1\% & 54.4\% & 69.9\% & 15.5\% \\ \hline
        FFX & 0.0\% & 38.1\% & 38.1\% & 42.5\% & 63.7\% & 21.2\% \\ \hline
        GP-GOMEA & 27.0\% & 46.9\% & 19.9\% & 90.7\% & 92.0\% & 1.3\% \\ \hline
        ITEA & 23.0\% & 39.8\% & 16.8\% & 40.7\% & 57.5\% & 16.8\% \\ \hline
        MRGP & 0.0\% & 38.1\% & 38.1\% & 98.2\% & 99.1\% & 0.9\% \\ \hline
        Operon & 17.7\% & 38.9\% & 21.2\% & 95.1\% & 94.7\% & -0.4\% \\ \hline
        SBP-GP & 12.4\% & 42.5\% & 30.1\% & 94.7\% & 95.6\% & 0.9\% \\ \hline
        gplearn & 16.4\% & 42.5\% & 26.1\% & 52.7\% & 65.5\% & 12.8\% \\ \hline
    \end{tabular}
\end{table}

\section{Effect of epoch count and number of training instances on GINN-LP performance}

\begin{figure*}[h]
    \centering
    \includegraphics[width=1\textwidth]{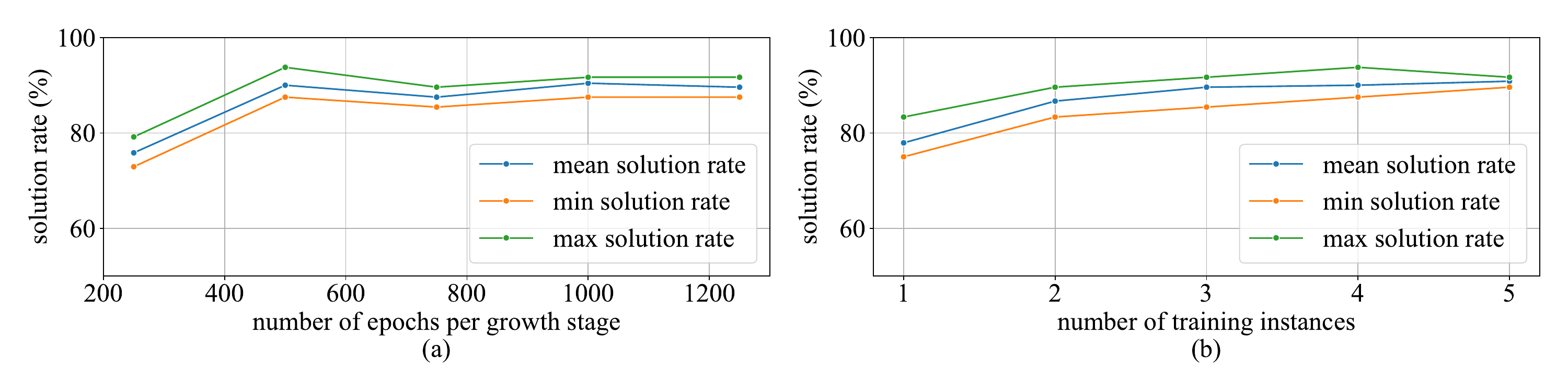}
    \caption{(a) Model performance with respect to the number of epochs per network growth stage. (b) Model performance (solution rate) with a varying number of training instances}
    \label{fig:results_epoch_inst}
\end{figure*}

\textbf{Effect of epoch count}. At each network growth stage, the network is trained for a pre-determined number of epochs. We train GINN-LP with a varying number of epochs per network growth stage (on all 48 datasets, 5 trials per dataset). As demonstrated in Figure \ref{fig:results_epoch_inst}(a), training each network growth stage for 500 epochs showed the highest performance and therefore was used for the rest of the experiments.

\textbf{Impact of the number of training instances}. We train multiple instances of GINN-LP and choose the best model as per Equation \ref{best_model}. The performance results for fixed training instance counts between 1 and 5 are shown in Figure \ref{fig:results_epoch_inst}(b). With more training instances, the likelihood of at least one recovering the correct ground-truth equation rises, increasing the overall solution rate. Due to increased training time with more instances, we opt for four instances, striking a balance between speed and performance.

\end{document}